# Deep learning-based super-resolution in coherent imaging systems

*Tairan Liu[1,2,3†], Kevin de Haan[1,2,3†], Yair Rivenson[1,2,3†], Zhensong Wei[1†], Xin Zeng[1], Yibo Zhang[1,2,3], and Aydogan Ozcan[1,2,3,4,*]*

[1]*Electrical and Computer Engineering Department, University of California, Los Angeles, CA, 90095, USA.*
[2]*Bioengineering Department, University of California, Los Angeles, CA, 90095, USA.*
[3]*California NanoSystems Institute (CNSI), University of California, Los Angeles, CA, 90095, USA.*
[4]*Department of Surgery, David Geffen School of Medicine, University of California, Los Angeles, CA, 90095, USA.*

*† Equally contributing authors*

*\* ozcan@ucla.edu*

**Abstract:** We present a deep learning framework based on a generative adversarial network (GAN) to perform super-resolution in coherent imaging systems. We demonstrate that this framework can enhance the resolution of both pixel size-limited and diffraction-limited coherent imaging systems. We experimentally validated the capabilities of this deep learning-based coherent imaging approach by super-resolving complex images acquired using a lensfree on-chip holographic microscope, the resolution of which was pixel size-limited. Using the same GAN-based approach, we also improved the resolution of a lens-based holographic imaging system that was limited in resolution by the numerical aperture of its objective lens. This deep learning-based super-resolution framework can be broadly applied to enhance the space-bandwidth product of coherent imaging systems using image data and convolutional neural networks, and provides a rapid, non-iterative method for solving inverse image reconstruction or enhancement problems in optics.

## 1. Introduction

Coherent imaging systems are advantageous for applications where the complex field information of the specimen is of interest [1]. Since Gabor's seminal work, various optical and numerical techniques have been suggested [2] to acquire the complex field of a coherently illuminated specimen. This has resulted in the characterization of both its absorption and scattering properties, as well as enabling numerical refocusing at different depths within a sample volume. To infer an object's complex field in a coherent optical imaging system, the "missing phase" should be retrieved. A classical solution to this missing phase problem is given by off-axis holography [3,4], which in general results in a reduction of the space-bandwidth product of the imaging system. In-line holographic imaging, which can be used to design compact microscopes [5], has utilized measurement diversity to generate a set of physical constraints for iterative phase retrieval [6–10]. Recently, deep-learning based holographic image reconstruction techniques have also been demonstrated to create a high-fidelity reconstruction from a single in-line hologram [11,12], with the possibility to further extend the depth-of-field of the reconstructed image [13].

Several approaches have been demonstrated to increase the resolution of coherent imaging systems [14–19]. Most of these techniques require sequential measurements and assume a quasi-static object while a set of diverse measurements are performed on the object, often using additional hardware, or sacrificing another degree of freedom such as the sample field-of-view [20]. In recent years, sparsity-based holographic reconstruction methods have also been demonstrated to increase the resolution of coherent imaging systems without the need for additional measurements or hardware [21–24]. Sparse signal recovery methods employed in coherent imaging are based on iterative optimization algorithms, which usually involve a comprehensive search over a parameter space to get the optimal object image and generally result in longer reconstruction times.

Recently, deep learning-based approaches have emerged to achieve super-resolution in incoherent microscopy modalities such as brightfield and fluorescence microscopy [25–29]. These data-driven super-resolution approaches produce a trained deep convolutional neural network that learns to transform low-resolution images into high-resolution images in a single feed-forward (i.e., non-iterative) step. In this paper we apply deep learning to enhance the resolution of coherent imaging systems and demonstrate a generative adversarial network (GAN) [30] that is trained to super-resolve both pixel-limited and diffraction-limited images. Furthermore, we demonstrate the success of this framework on biomedical samples such as thin sections of lung tissue and Papanicolaou (Pap) smear samples. We quantify our results using the structural similarity index (SSIM) [31] and spatial frequency content of the network's output images in comparison to the higher resolution images (which constitute our ground truth). This data-driven image super-resolution framework is applicable to enhance the performance of various coherent imaging systems.

## 2. Methods

The raw holograms were collected using an in-line holographic imaging geometry implemented in two different set-ups as illustrated in Fig. 1a and Fig. 1b for pixel size-limited and diffraction-limited coherent microscopy, respectively. The network's training and testing images were generated by processing these raw holograms with an iterative multi-height phase recovery algorithm [5,7,32–35]. Using this method, a high- and low-resolution image pair for each sample was reconstructed. These images were in turn finely registered to each other and split into small patches for training. For the pixel-super-resolution network, which aims to super-resolve a pixel size-limited image, the real and imaginary components of the phase recovered image pairs were used to train the network (summarized in Fig. 2). The second super-resolution network is designed to improve the resolution of diffraction-limited coherent images. In this case only the phase channel was used to train the network (summarized in Fig. 3).

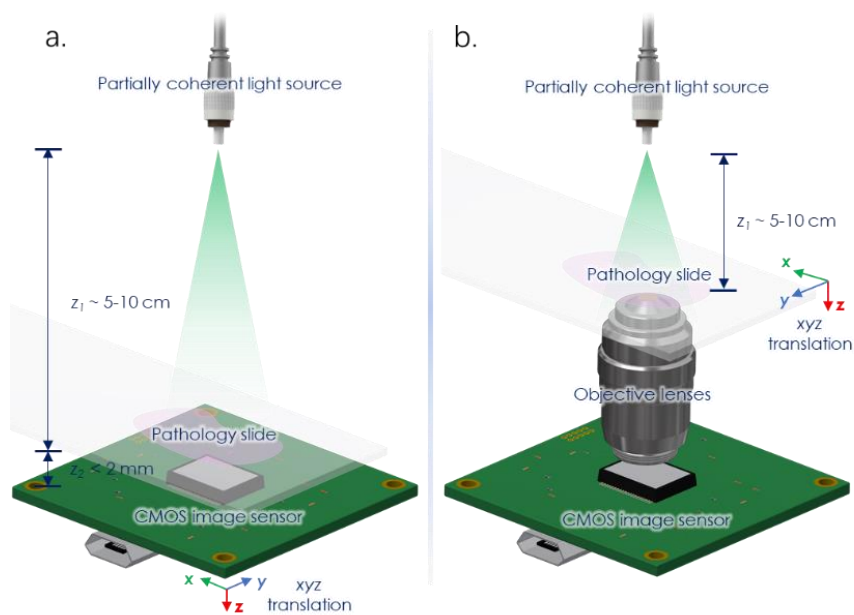

Fig. 1. Schematic of the coherent imaging systems used in this work. (a) A Lens-free on-chip holographic microscope. The sample is placed at a short distance ($z_2 < 2$ mm) above the image sensor chip. The resolution of this lensless on-chip imaging modality (without the use of additional degrees of freedom) is pixel size-limited due to its unit magnification. (b) A lens-based in-line holographic microscope, implemented by removing the condenser and switching the illumination to a partially-coherent light source on a conventional bright-field microscope. The resolution in this case is limited by the NA of the objective lens.

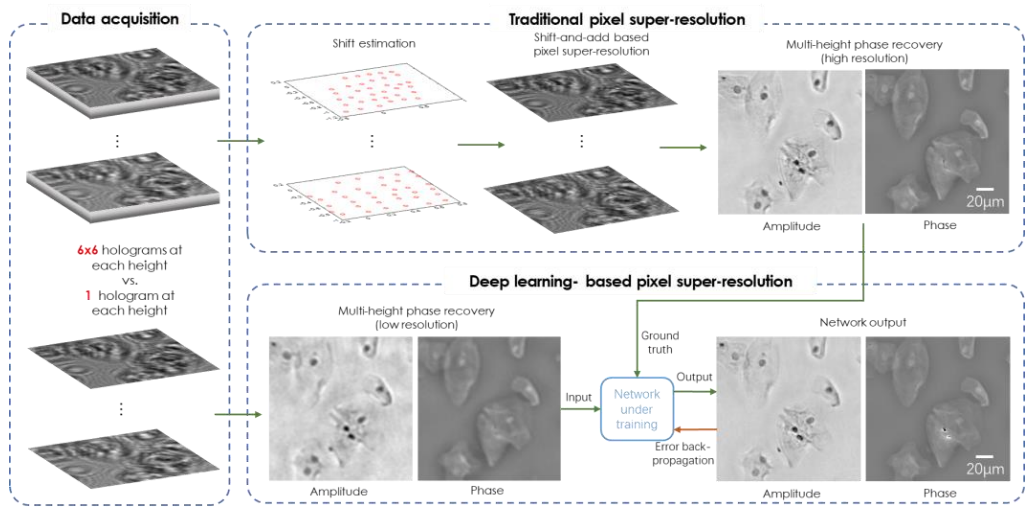

Fig. 2. Schematic of the training process for deep-learning based pixel super-resolution.

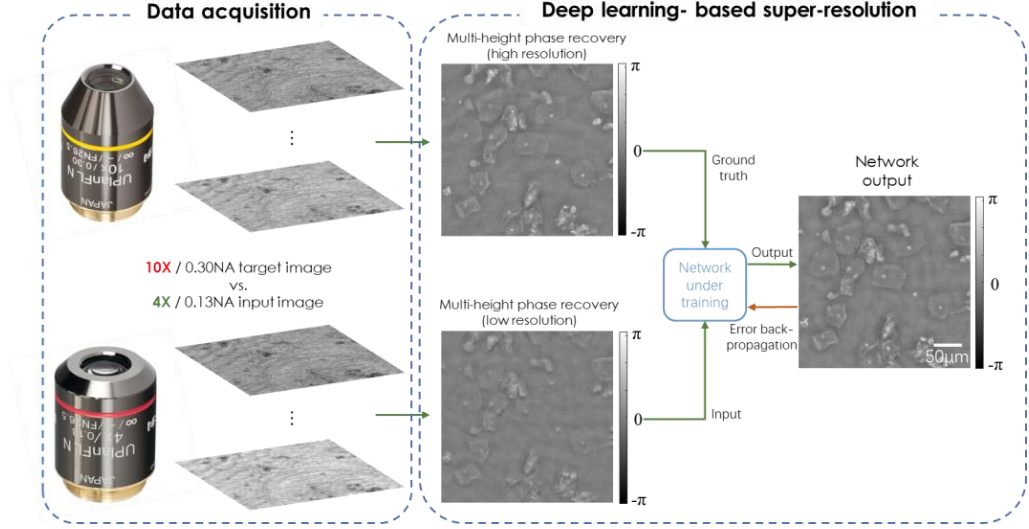

Fig. 3. Schematic of the training process for deep learning-based optical super-resolution for an NA-limited coherent imaging system.

## 2.1 Generation of ground truth super-resolved image labels

For the pixel size-limited coherent imaging system (Fig. 1a), the ground truth super-resolved images were created by collecting multiple lower resolution holograms at different lateral positions, where the CMOS image sensor was shifted by a mechanical stage using sub-pixel shifting. Once an accurate shift table was estimated, a shift-and-add based pixel super-resolution algorithm [32] was applied. In our set-up, we used an illumination wavelength of 550 nm with a bandwidth ($\Delta\lambda$) of ~2 nm (WhiteLase Micro with acousto-optic tunable filter, NKT Photonics), a single mode fiber (QPMJ-3S2.5A-488-3.5/125-1-0.3-1) with a core diameter of ~3.5 μm and a source-to-sample distance ($z_1$) of ~5 cm. As a result, the effective spatial coherence diameter at the sensor plane is larger than the width of the CMOS imager chip used in our on-chip imaging system. Therefore, the achievable resolution is limited by the temporal coherence length of the illumination [36], which is defined by:

$$\Delta L_c \approx \sqrt{\frac{2\ln 2}{\pi n}} \cdot \frac{\lambda^2}{\Delta\lambda} = 100.47 \ \mu m \tag{1}$$

Then, assuming a sample-to-sensor distance ($z_2$) of ~300 μm, the effective numerical aperture (NA) of our set-up will be limited by the temporal coherence of the source, and can be estimated as:

$$\mathrm{NA} = n\sin\theta = n\sqrt{1-\cos^2\theta} = n\sqrt{1-\left(\frac{z_2}{z_2 + \Delta L_c}\right)^2} \approx 0.6624 \tag{2}$$

Based on this effective numerical aperture, ignoring the pixel size at the hologram plane, the achievable coherence-limited resolution of our on-chip microscope can be approximated as [4]:

$$d \propto \frac{\lambda}{\mathrm{NA}} = \frac{0.55}{0.6624} = 0.8303 \ \mu m \tag{3}$$

At the hologram/detector plane, however, the effective pixel pitch of the CMOS image sensor (IMX 081, Sony RGB sensor, pixel size of 1.12 μm) using only one color channel is 2.24 μm. Based on this, the effective pixel size for each ground truth image after the application of the pixel super-resolution algorithm to 4 raw holograms (2×2 lateral positions), 9 raw holograms (3×3 lateral positions), and 36 raw holograms (6×6 lateral positions) would be 1.12 μm, 0.7467 μm and 0.3733 μm, respectively. Based on [Eq. (3)], the effective pixel size achieved by pixel super-resolution using 6×6 lateral positions can adequately sample the specimen's holographic diffraction pattern and is limited by temporal coherence. All of the other images (using 1×1, 2×2 and 3×3 raw holograms) remain pixel-limited in their achievable spatial resolution. This pixel-limited resolution of an on-chip holographic microscope is in general a result of its unit magnification, which achieves a large imaging field-of-view (FOV) that is only limited by the active area of the opto-electronic image sensor chip. This can easily reach 20-30 mm$^2$ and >10 cm$^2$ using state-of-the-art CMOS and CCD imagers, respectively [5].

For the second set-up, which uses lens-based holographic imaging for diffraction-limited coherent microscopy, the lower and higher resolution images were acquired by using different objective lenses. In this set-up, the illumination was performed by a fiber coupled laser diode at an illumination wavelength of 532 nm. A 4×/0.13NA objective lens was used to acquire lower resolution images, achieving a diffraction limited resolution of ~4.09 μm and an effective pixel size of ~1.625 μm. A 10×/0.30NA objective lens was used to acquire the higher resolution images (ground truth labels), achieving a resolution of 1.773 μm and an effective pixel size of ~0.65 μm.

*2.2 Autofocusing and singular value decomposition-based background subtraction*

We used the free space angular spectrum propagation approach [4] as the building block of both the autofocusing and the multi-height phase recovery techniques used for reconstructing the holograms by both the pixel size-limited and diffraction-limited coherent imaging systems. In addition, after the phase recovery, all the complex-valued images in both types of coherent imaging systems were propagated to the sample plane. To perform multi-height phase recovery [5,7,32–35], holograms at 8 different sample-to-sensor distances were collected for both types of coherent imaging systems. For the lensfree holographic on-chip microscope, these heights were accurately estimated by using the Tamura of the gradient (ToG) edge sparsity criterion-based autofocusing algorithm [37]. However, for the lens-based diffraction-limited coherent imaging system, this autofocusing algorithm required a background subtraction step. For undesired particles or dust associated with the objective lens or other parts of the optical microscope, the diffraction pattern that is formed is independent of the sample and its position. Using this information, a singular value decomposition (SVD)-based background subtraction was performed [38], after which the ToG-based autofocusing algorithm was successfully applied.

*2.3 Multi-height phase recovery*

An iterative multi-height phase recovery technique [33] was applied to eliminate the twin image artifacts in both of the coherent imaging systems that were used in this work. To perform this, an initial zero-phase was assigned to the intensity/amplitude measurement at the 1st hologram height. Next, the iterative algorithm begins by propagating the complex field to each hologram height until the 8$^{th}$ height is reached, and then backpropagates the resulting fields until the 1$^{st}$ height is reached. While the phase was retained at each hologram height, the amplitude was updated by averaging current amplitude and the square root of the measured intensity at each height.

*2.4 Registration between lower resolution and higher resolution (ground truth) images*

Image registration plays a key role in generating the training and testing image pairs for the network in both the pixel size-limited and diffraction-limited coherent imaging systems. A pixel-wise registration must be performed to ensure the success of the network in learning the transformation to perform super-resolution.

For the pixel size-limited system, the low-resolution input images were bicubically up-sampled, and a correlation-based registration was performed. This framework corrects any rotational misalignments or shifts between the images. This registration process correlates the spatial patterns of the phase images and uses the correlation to establish an affine transform matrix. This can in turn be applied to the high-resolution images to ensure proper matching of the corresponding fields-of-view between the low-resolution images and their corresponding ground truth labels. Finally, each image is cropped by 50 pixels on each side to accommodate for any relative shift that may have occurred.

For the diffraction-limited coherent imaging system, an additional rough FOV matching step is required before the registration. For this step, the higher resolution phase images are first stitched together, by calculating the overlap between neighboring images, and then fusing them together into a larger image. The corresponding lower resolution phase images are then matched to this larger image, which is done by first extracting both the strong and weak edges of the image using the Canny edge detection method [39] with two separate thresholds. These edges of the up-sampled lower resolution image are then correlated with each patch of the large image. Whichever patch has the highest correlation is cropped out and is used as the input for the network.

## 2.5 GAN architecture and training process

Once the higher and lower resolution image pairs are accurately registered, they are cropped into small image patches (128×128 pixels), which are used to train the network. The GAN is made up of two separate networks which can be seen in Fig. 4. A generator ($G$) network is used to generate an image that has the same features as the label (ground truth) image, while the discriminator ($D$) tries to distinguish between the generated and label (ground truth) images. For both the pixel-size limited and the diffraction-limited coherent imaging systems, the discriminator loss function is defined as:

$$l_{\text{discriminator}} = D\big(G(x_{\text{input}})\big)^2 + \big(1 - D(z_{\text{label}})\big)^2 \qquad (4)$$

where $D(.)$ and $G(.)$ refer to the discriminator and generator network operators, respectively, $x_{\text{input}}$ is the lower resolution input to the generator, and $z_{\text{label}}$ is the higher resolution label image.

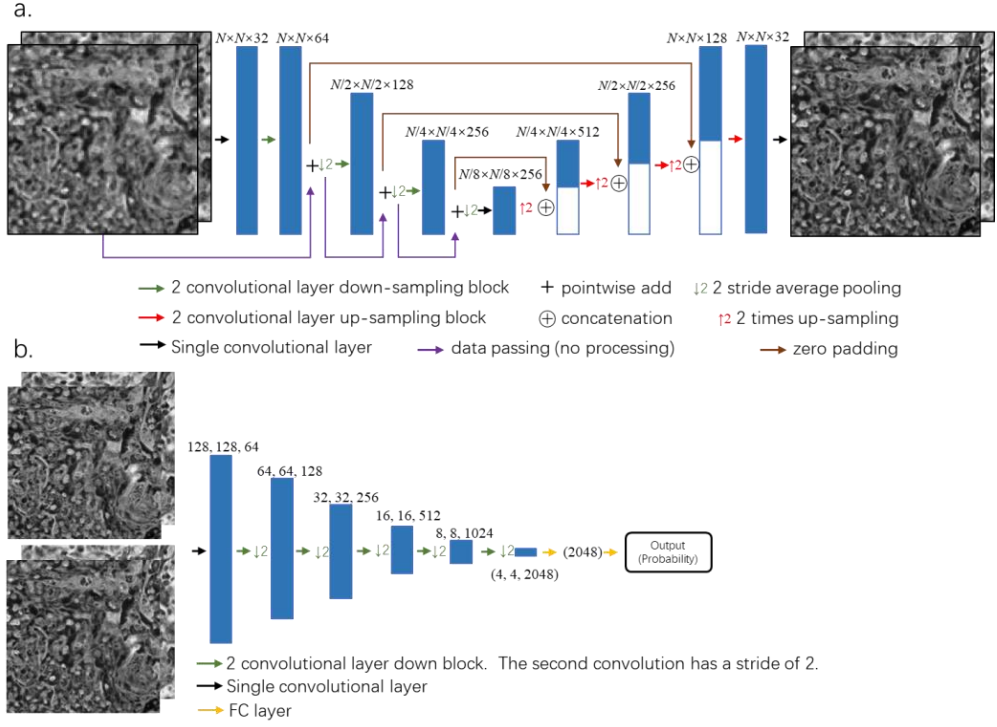

Fig. 4. Diagram of GAN structure. (a) Structure of the generator portion of the network. (b) Structure of the discriminator portion of the network.

For the lensfree holographic imaging system, the generator loss function is defined by:

$$l_{generator} = L_1\{z_{label}, G(x_{input})\} + \lambda \times TV\{G(x_{input})\} + \alpha \times \left(1 - D(G(x_{input}))\right)^2 \qquad (5)$$

The $L_1\{z_{label}, G(x_{input})\}$ term is calculated using:

$$\mathrm{E}_{n\_pixels}\left(\mathrm{E}_{n\_channels}\left(|G(x_{input}) - z_{label}|\right)\right) \qquad (6)$$

This finds the absolute difference between each pixel of the generator output image and its corresponding label. $\mathrm{E}_{n\_pixels}(.)$ and $\mathrm{E}_{n\_channels}(.)$ are the expectation values for the pixels within each image and the channels of each image, respectively. $TV\{G(x_{input})\}$ represents the total variation loss, which acts as a regularization term, applied to the generator output. The total variation ($TV$) is calculated with the following equation:

$$TV = \sum_{i,j}\left|G(x_{input})_{i+1,j} - G(x_{input})_{i,j}\right| + \left|G(x_{input})_{i,j+1} - G(x_{input})_{i,j}\right| \qquad (7)$$

The $i$ and $j$ indices represent the location of the pixels within the image.

The last term in [Eq. (5)] ($\alpha \times \left(1 - D(G(x_{input}))\right)^2$) is a function of how well the output image of the generator network can be predicted by the discriminator network. $\alpha$ and $\lambda$ are regularization parameters which were set to 0.00275 and 0.015 respectively. The $L_1$ loss term, $L_1\{z_{label}, G(x_{input})\}$, made up 60% of the overall loss, while the total variation term, $\lambda \times TV\{G(x_{input})\}$, was approximately 0.25% of the total loss. The discriminator loss term, $\alpha \times (1-D(G(x_{input})))^2$, made up the remainder of the overall generator loss. Once the networks have been successfully trained we reach a state of equilibrium where the discriminator network cannot successfully discriminate between the output and label images, and $D(G(x_{\text{input}}))$ converges to approximately 0.5.

The loss function for the lens-based coherent microscope incorporates an additional structural similarity index (SSIM) [31] term in addition to the terms included for the lensfree on-chip imaging system, i.e.,:

$$l_{\text{generator}} = L_1\left\{z_{\text{label}}, G(x_{\text{input}})\right\} + \lambda \times TV\left\{G(x_{\text{input}})\right\} + \alpha \times \left(1 - D(G(x_{\text{input}}))\right)^2 \\ + \beta \times \text{SSIM}\left\{z_{\text{label}}, G(x_{\text{input}})\right\} \quad (8)$$

where SSIM{z, x} is defined as [31]:

$$\text{SSIM}(x, y) = \frac{(2\mu_z\mu_x + c_1)(2\sigma_{z,x} + c_2)}{(\mu_z^2 + \mu_x^2 + c_1)(\sigma_z^2 + \sigma_x^2 + c_2)} \quad (9)$$

where $\mu_z$, $\mu_x$ are the averages of $z$, $x$; $\sigma_z^2, \sigma_x^2$ are the variances of $z$, $x$, respectively; $\sigma_{z,x}$ is the covariance of $z$ and $x$; and $c_1$, $c_2$ are dummy variables used to stabilize the division with a small denominator. The term $\beta\times\text{SSIM}\{z_{\text{label}},G(x_{\text{input}})\}$ is set to make up ~15% of the total generator loss, with the rest of the regularization weights reduced in value accordingly.

Our generator network uses an adapted U-net architecture [40]. The network begins with a convolutional layer that increases the number of channels to 32 and a leaky rectified linear (LReLU) unit, defined as:

$$\text{LReLU}(x) = \begin{cases} x & \text{for } x > 0 \\ 0.1x & \text{otherwise} \end{cases} \quad (10)$$

Following this layer, there is a down-sampling and an up-sampling section. Each section consists of three distinct layers, each made up of separate convolution blocks (see Fig. 4a). For the down-sampling section, these residual blocks consist of two convolution layers with LReLU units acting on them. At the output of the second convolution of each block the number of channels is doubled. The down-sampling blocks are connected by an average-pooling layer of stride two that down-samples the output of the previous block by a factor of two in both lateral dimensions (see Fig. 4a).

The up-sampling section of the network uses a reverse structure to reduce the number of channels and return each channel to its original size. Similar to the down sampling section, each block contains two convolutional layers, using LReLU. At the input of each block, the previous output is up-sampled using a bilinear interpolation and concatenated with the output of the down-sampling path at the same level (see Fig. 4a). Between the two paths, there is a convolutional layer which maintains the number of the feature maps from the output of the last residual block to the beginning of the down-sampling path (Fig. 4a). Finally, a convolutional layer reduces the number of output channels to match the size of the label.

The discriminator portion of the network is made up of a convolutional layer, followed by five discriminator blocks, an average pooling layer and two fully connected layers, which reduce the output to a single value (see Fig. 4b). Both the label images and the output of the generator network are input into the discriminator network. These images are input into the initial convolutional layer, which is used to increase the number of channels to 32. The five discriminator blocks all contain two convolutional layers with LReLU functions. The first convolution maintains the size of the output, and the second doubles the number of channels while halving the size of the output in each lateral dimension. Next, the average pooling layer is used to find the mean of each channel, reducing the dimensionality to a vector of length 1024 for each patch. Each of these vectors is subsequently fed into two fully connected layers and LReLU functions in series. While the first fully connected layer does not change the dimensionality, the second reduces the output of each patch to a single number which is input into a sigmoid function. The output of the sigmoid function represents the probability of the input being either real or fake and is used as part of the generator's loss function.

The filter size for each convolution was set to be 3×3. The trainable parameters are updated using an adaptive moment estimation (Adam) [41] optimizer with a learning rate $1\times10^{-4}$ for the generator network and $1\times10^{-5}$ for the discriminator network. The image data were augmented by randomly flipping 50% of the images, and randomly choosing a rotation angle (0, 90, 180, 270 degrees). For each iteration that the discriminator is updated, the generator network is updated four times, which helps the discriminator avoid overfitting to the target images. The convolutional layer weights are initialized using a truncated normal distribution while the network bias terms are initialized to zero. A batch size of 10 is used for the training, and a batch size of 25 is used for validation. The networks chosen for blind testing were those with the lowest validation loss.

*2.6 Software implementation details*

The network was developed using a desktop computer running the Windows 10 operating system. The desktop uses an Nvidia GTX 1080 Ti GPU, a Core i7-7900K CPU running at 3.3 GHz, and 64 GB of RAM. The network was programmed using Python (version 3.6.0) with the TensorFlow library (version 1.7.0). The number of training steps as well as the training time for each network are reported in Table 1, and the testing times are reported in Table 2.

**Table 1. Training details for the deep neural networks.**

| Resolution limiting factor | Tissue type | Low resolution input type | Training time (s) | Number of iterations |
|---|---|---|---|---|
| Diffraction-limited | Pap smear | 4×/0.13 NA objective lens | 46411 | 100000 |
| Pixel size-limited | Pap smear | 1×1 raw hologram | 9078 | 17000 |
| | Lung | 1×1 raw hologram | 17052 | 28000 |
| | Lung | 2×2 raw holograms | 9363 | 15000 |
| | Lung | 3×3 raw holograms | 30480 | 52500 |

All the networks were trained with a batch size of 10 using 128×128 pixel patches.

**Table 2. Time for each network to output a 1940×1940 pixel image.**

| Resolution limiting factor | Tissue type | Low resolution input type | Testing Time (s/image) |
|---|---|---|---|
| Diffraction-limited | Pap smear | 4×/0.13 NA objective lens | 1.26 |
| Pixel size-limited | Pap smear | 1×1 raw hologram | 1.42 |
| | Lung | 1×1 raw hologram | 1.37 |
| | Lung | 2×2 raw holograms | 1.38 |
| | Lung | 3×3 raw holograms | 1.38 |

Each measurement is the average time, calculated using 150 test images.

## 3. Results and Discussion

*3.1 Super-resolution of a pixel size-limited coherent imaging system*

We first report the performance of the network for the pixel size-limited coherent imaging system using a Pap smear sample and a Masson's trichrome stained lung tissue section (connected tissue sample). For the Pap smear, two samples from different patients were used for training. For the lung tissue samples, three tissue sections from different patients were used for training. The networks were blindly tested on additional tissue sections from other patients. The FOV of each tissue image was ~20 mm$^2$ (corresponding to the sensor active area).

Fig. 5 illustrates the network's super-resolved output images along with pixel-size limited lower resolution input images and the higher resolution ground truth images of a Pap smear sample. The input images have a pixel pitch of 2.24 μm, and the label images have an effective pixel size of 0.37 μm (see the Methods section). For lung tissue sections, we also demonstrate our super-resolution results (Fig. 6) using three different deep networks, where the input images for each network has a different pixel size (i.e., 2.24 μm, 1.12 μm, and 0.7467 μm, corresponding to 1×1, 2×2 and 3×3 lateral shifts, respectively, as detailed in the Methods section). In comparison to the less densely connected Pap smear sample results, the network output misses some spatial details for lung tissue imaging when the input pixel size is at the coarsest level of 2.24 μm. These spatial features/details are recovered back by the other two networks that use smaller input pixels as shown in Fig. 6.

We also report the SSIM values with respect to the reference label images in order to further evaluate the performance of our network output when applied to a pixel size-limited coherent imaging system. The average SSIM values for the entire image FOV (~20 mm$^2$) are listed in Table 3, where the input SSIM values were calculated between the bicubic interpolated lower resolution input images and the ground truth images. The results clearly demonstrate the improved structural similarity of the network output images.

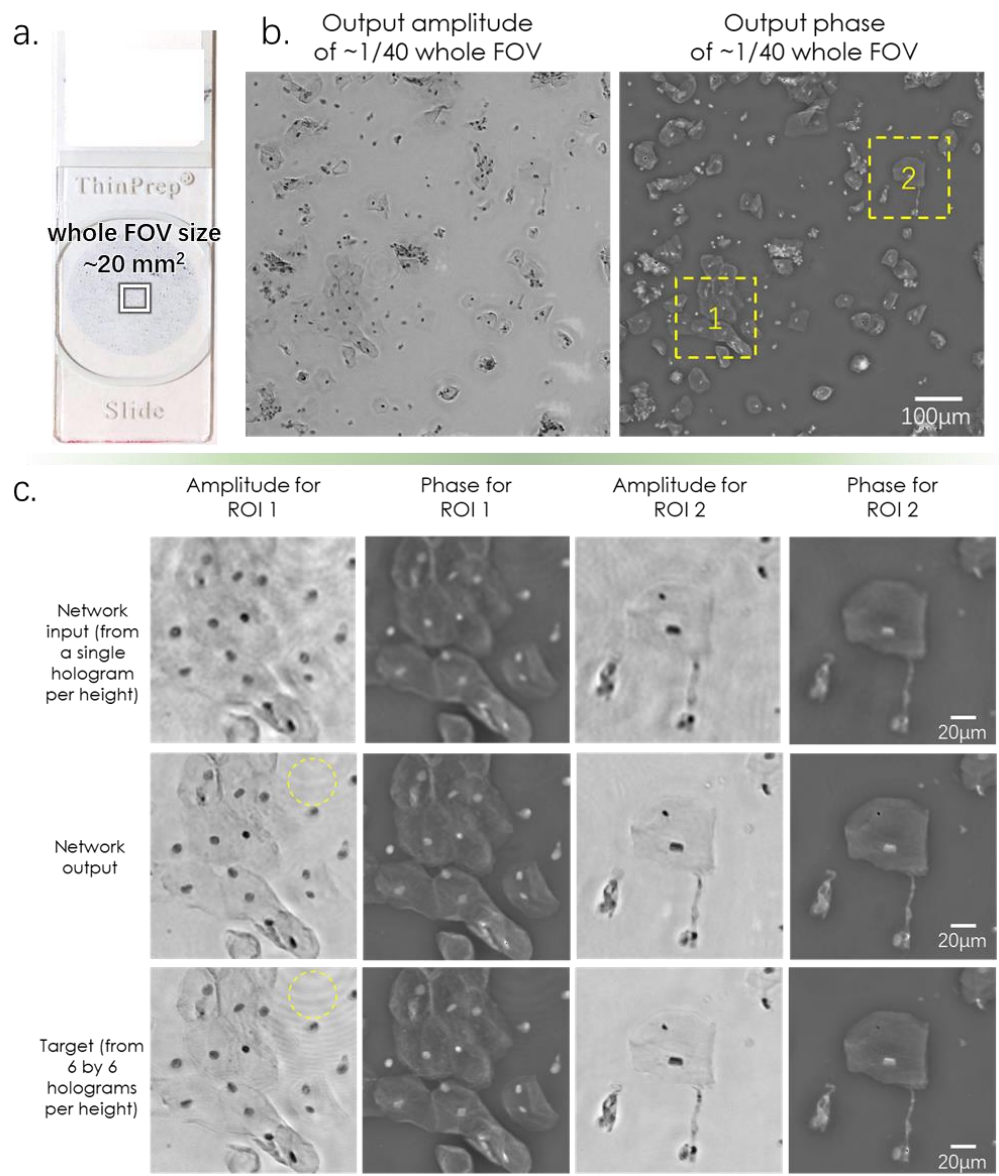

Fig. 5. Deep learning-based pixel super-resolution imaging of a Pap smear slide under 550 nm illumination. (a) Whole FOV of the lensfree imaging system. (b) Amplitude and phase channels of the network output. (c) Further zoom-in of (b) for two regions of interest. The marked region in the first column demonstrates the network's ability to process the artifacts caused by out-of-focus particles within the sample.

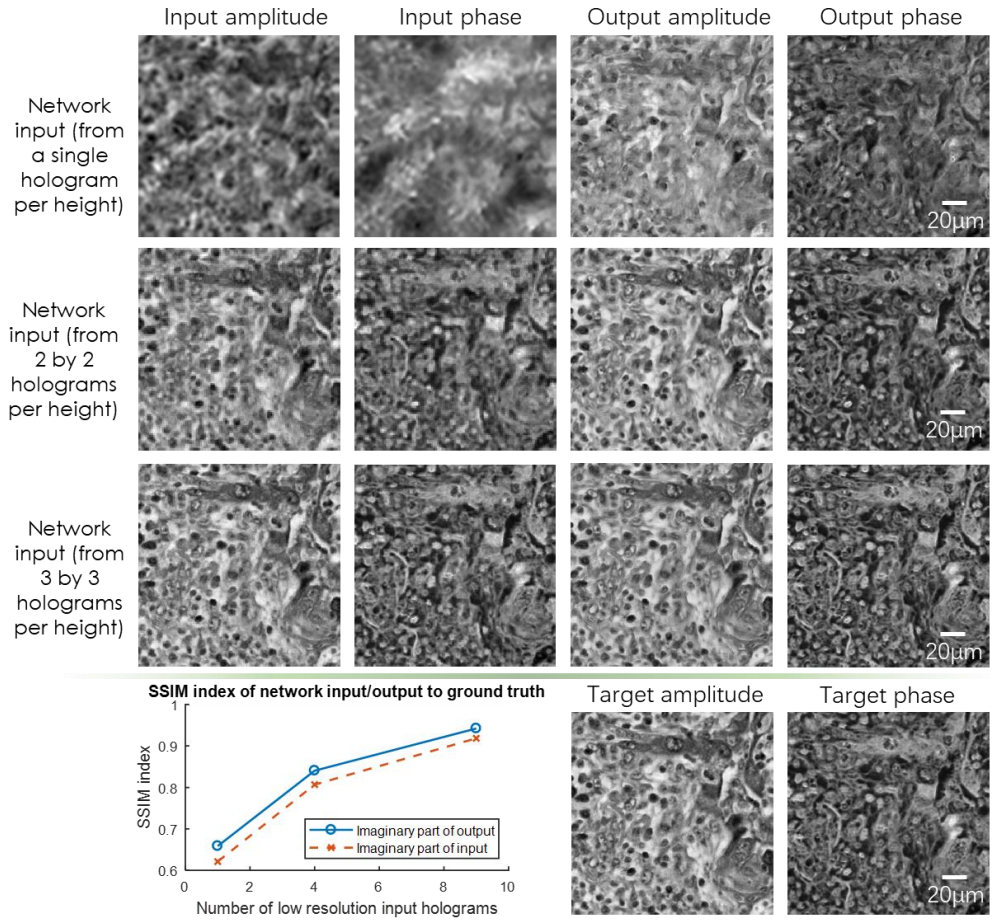

Fig. 6. Comparison of the performances for the deep-learning-based pixel super-resolution methods using different input images. The sample is a Masson's trichrome stained lung tissue slide, imaged at an illumination wavelength of 550 nm. SSIM values are also shown for the network input and output images for each case. The ground truth (target) image for each SSIM value is acquired using 6×6 lensfree holograms per height.

Table 3: Average SSIM values for the lung and Pap smear samples for the deep neural network output (also see Figs. 5 and 6 for sample images in each category).

| Resolution limiting factor | Tissue type | Low resolution input type | Input SSIM | | Output SSIM | |
|---|---|---|---|---|---|---|
| | | | Imaginary | Real | Imaginary | Real |
| Pixel size-limited | Pap smear | 1×1 raw hologram | 0.9097 | 0.9135 | 0.9392 | 0.9442 |
| | Lung | 1×1 raw hologram | 0.6213 | 0.5404 | 0.6587 | 0.7135 |
| | Lung | 2×2 raw holograms | 0.8069 | 0.8205 | 0.8405 | 0.8438 |
| | Lung | 3×3 raw holograms | 0.9185 | 0.9184 | 0.9422 | 0.9347 |

In addition to SSIM comparison, we also report the improved performance of our network output using spatial frequency analysis: Fig. 7 reports the 2-D spatial frequency spectra and the associated radially-averaged frequency intensity of the network input, network output and the ground truth images corresponding to our lensfree on-chip imaging system. The appearance of the higher spatial frequency components in the output of the network, approaching to the spatial frequencies of the ground truth image is another indication of our super-resolution performance.

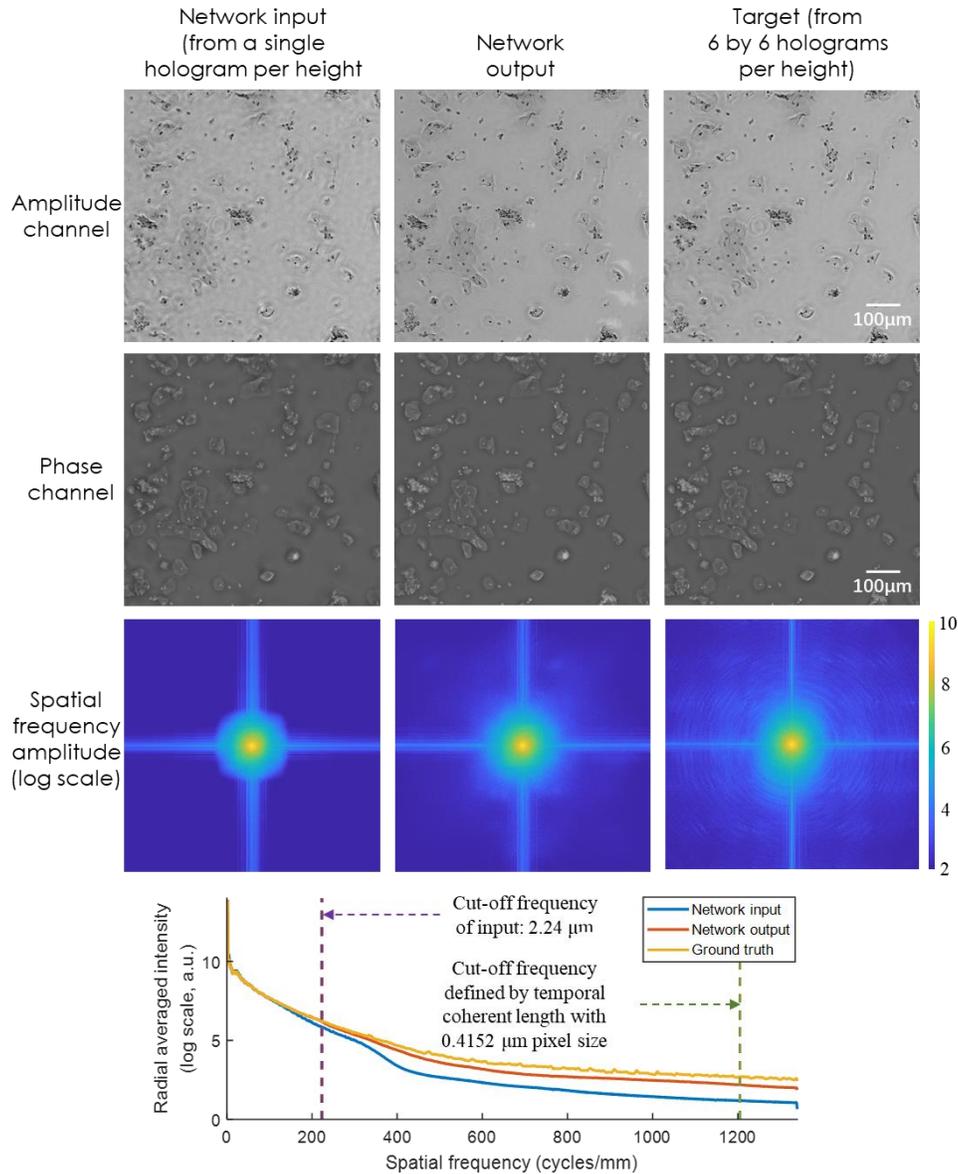

Fig. 7. Radially-averaged spatial frequency spectra of the network input, network output and target images, corresponding to a lensfree on-chip coherent imaging system.

*3.2 Super-resolution of a diffraction-limited coherent imaging system*

For the task of super-resolving a diffraction-limited coherent imaging system, we trained an identical network architecture (see the Methods section) with a Pap smear sample. As in the pixel super-resolution case reported earlier, two samples were obtained from two different patients, and the trained network was blindly tested on a third sample obtained from a third patient. The input images were obtained using a 4×/0.13 NA objective lens and the reference ground truth images were obtained by using a 10×/0.30 NA objective lens. Fig. 8 illustrates a visual comparison of the network input, output and label images, providing the same conclusions as in Fig. 5 and Fig. 6. Similar to the pixel size-limited coherent imaging system, we also analyzed the performance of our network using spatial frequency analysis which is reported in Fig. 9. The higher spatial frequencies of the network output image approach the spatial frequencies observed in the ground truth images, similar to the results of Fig. 7.

On the other hand, the SSIM criterion did not reveal the same trend that we observed in the lensfree on-chip microscopy system reported earlier, and only showed a very small increase from e.g., 0.876 for the input image to 0.879 for the network output image. This is mostly due to increased coherence related artifacts and noise, compared to the lensfree on-chip imaging set-up, since the lens-based design has several optical components and surfaces within the optical beam path, making it susceptible to coherence induced background noise and related image artifacts, which partially dominate SSIM calculations.

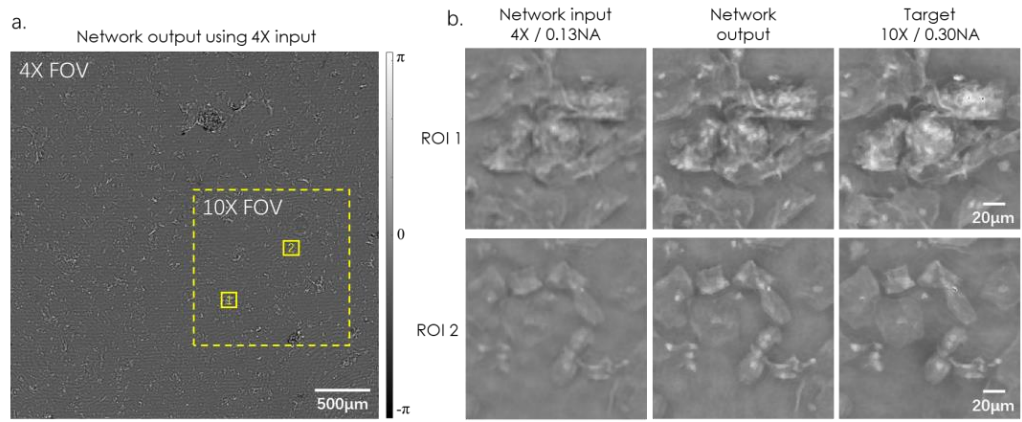

Fig. 8. Deep learning-based super-resolution imaging of a Pap smear slide under 532 nm illumination using a lens-based holographic microscope. (a) Phase channel of the network output image. (b) Zoomed-in images of (a).

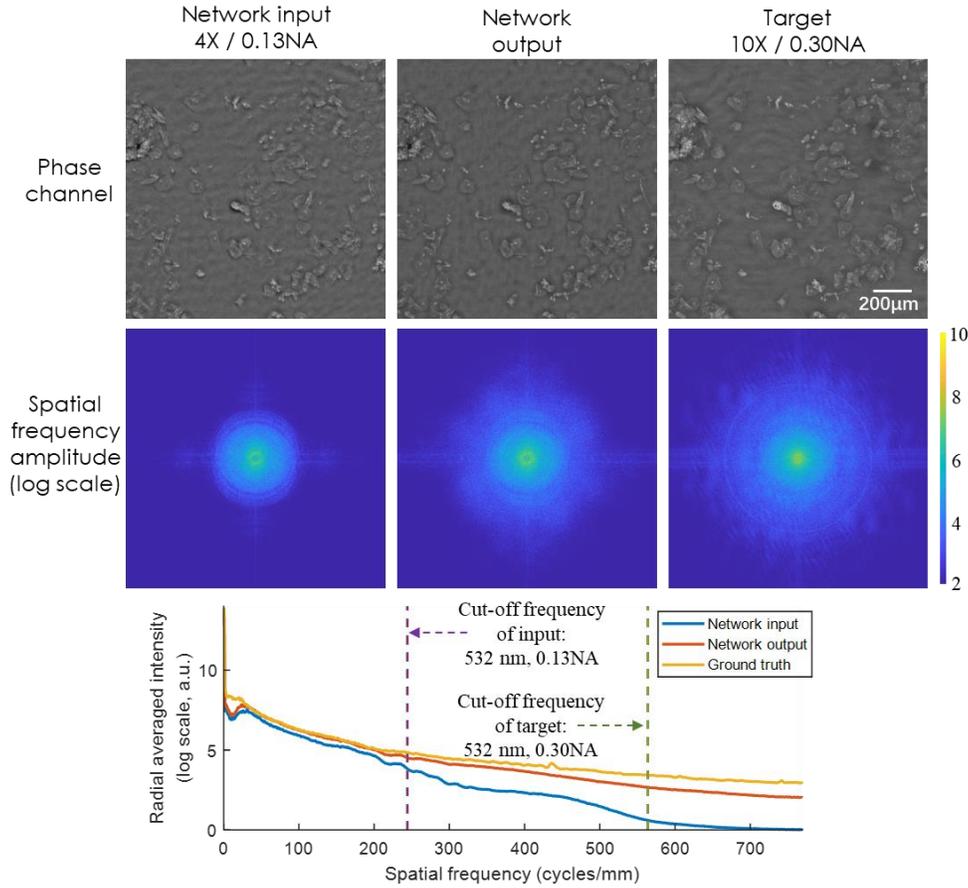

Fig. 9. Radially-averaged spatial frequency spectra of the network input, network output and target images, corresponding to a lens-based coherent imaging system.

## 4. Conclusion

We have presented a GAN-based framework for super-resolution of pixel size limited and diffraction limited coherent imaging systems. The framework was demonstrated on biologically connected thin tissue sections (lung and Pap smear samples) and the results were quantified using structural similarity index and spatial frequency spectra analysis. The proposed framework provides a highly optimized, non-iterative reconstruction engine that rapidly produces resolution enhancement, without additional parameter optimization.

The proposed approach is not restricted to a specific coherent imaging modality and is broadly applicable to various coherent image formation techniques. One of the techniques that will highly benefit from the proposed framework is off-axis holography. The proposed technique might be used to bridge the space-bandwidth-product gap between off-axis and in-line coherent imaging systems, while retaining the single-shot and high sensitivity advantages of off-axis image acquisition systems.